\documentclass{article}
\usepackage{ijcai19}

\usepackage{times}
\usepackage{soul}
\usepackage{url}
\usepackage[hidelinks]{hyperref}
\usepackage[utf8]{inputenc}
\usepackage[small]{caption}
\usepackage{graphicx}
\usepackage{amsmath}
\usepackage{booktabs}
\usepackage{array}
\usepackage{multirow}
\usepackage{multicol}
\usepackage{tabu}
\usepackage{times}
\usepackage{epsfig}
\usepackage{graphicx}
\usepackage{amsmath}
\usepackage{amssymb}
\usepackage{amsfonts}
\usepackage{color}
\usepackage{colortbl}
\usepackage{epsfig}
\usepackage{multirow}
\usepackage{algorithm}
\usepackage{subcaption}
\usepackage{algpseudocode}
\usepackage{scrextend}
\usepackage{tabularx}
\usepackage{bbm}
\usepackage{breqn}

\title{Fast Adaptation of Deep Models for Facial Action Unit Detection\\ Using Model-Agnostic Meta-Learning}

\author{
Mihee Lee$^1$\and
Ognjen (Oggi) Rudovic$^{2,3}$\and
Vladimir Pavlovic$^{1,4}$\And
Maja Pantic$^{4,5}$\\
\affiliations
$^1$Rutgers University, Piscataway, NJ, USA\\
$^2$MIT Media Lab, Cambridge, MA, USA\\
$^3$Augsburg University, Germany\\
$^4$Samsung AI Center, Cambridge, UK\\
$^5$Imperial College, London, UK\\
}

\begin{document}

\maketitle

\begin{abstract}
Detecting facial action units (AU) is one of the fundamental steps in automatic recognition of facial expression of emotions and cognitive states. Though there have been a variety of approaches proposed for this task, most of these models are trained only for the specific target AUs, and as such they fail to easily adapt to the task of recognition of new AUs (i.e., those not initially used to train the target models). In this paper, we propose a deep learning approach for facial AU detection that can easily and in a fast manner adapt to a new AU or target subject by leveraging only a few labeled samples from the new task (either an AU or subject). To this end, we propose a modelling approach based on the notion of the model-agnostic meta-learning~\cite{Finn2017ICML}, originally proposed for the general image recognition/detection tasks (e.g., the character recognition from the Omniglot dataset). Specifically, each subject and/or AU is treated as a new learning task and the model learns to adapt based on the knowledge of the previous tasks (the AUs and subjects used to pre-train the target models). Thus, given a new subject or AU, this meta-knowledge (that is shared among training and test tasks) is used to adapt the model to the new task using the notion of deep learning and model-agnostic meta-learning. We show on two benchmark datasets (BP4D and DISFA) for facial AU detection that the proposed approach can be easily adapted to new tasks (AUs/subjects). Using only a few labeled examples from these tasks, the model achieves large improvements over the baselines (i.e., non-adapted models).
\end{abstract}

\section{Introduction}
 Facial Action Units (AUs) represent a set of facial muscle activation, where a number of AUs (more than 36) are defined by Facial Action Coding System (FACS). Based on the combinations of AUs, any possible facial expression can be described and further interpretations of, e.g., emotional and cognitive states be made. For this reason, the detection of facial AUs from face images is an important task in computer vision. However, the manual annotation of AUs is difficult, time-consuming and tedious task. Furthermore, to perform an objective and consistent labeling of AUs, the human experts with comprehensive understanding of AUs are needed, which is costly. Moreover, manual annotation of AUs does not scale well with large image datasets of human facial expressions, typically required for automated systems for facial expression analysis. Though there are several benchmark datasets with annotated AUs, such as DISFA \cite{Mavadati2013IEEE}, BP4D \cite{Mavadati2013IEEE}, CK+ \cite{Lucey2010IEEE}, and MMI \cite{Pantic2005IEEE}, the ultimate challenge arrives when faced with face images of new subjects and/or AUs that we wish to detect from their face images. Usually, an AU detection model is trained for a number of available AU annotations. However, these models do not generalize well when tested on new tasks such as new subjects nor can easily be adapted to new AUs, without a substantial number of ground truth labels. For instance, to tackle this, several unsupervised re-learning approaches (e.g., ~\cite{Wen2017tpami,Can2018ACM}) transferred the salient information of a new subject to the trained model in order to have better prediction for a new subject. However, these approaches do not use any labeled data from the new task (subject), and they typically require a lot of data to adapt to a new task. Yet, in the real-life situations, the models that can easily and in a fast way to the new tasks are required. Another challenge for existing AU detection models is the imbalanced AU distribution. Many AUs have significantly more negative (no AU activation) than positive (where the target AU is active) training examples. This easily leads to highly biased models. More importantly, when the model needs be fine-tuned very quickly, its adaptation is even more challenging due to this inherent imbalanced in the AU data.  

 
To tackle some of these challenges, this paper introduces a deep learning approach for AU detection that can easily be adapted to new tasks (AUs and/or subjects) using the notion of model-agnostic meta-learning~\cite{Finn2017ICML}. This approach is based on the few-shot learning paradigm, designed to perform an efficient model adaptation using only a few labelled examples of target tasks. This, in turn, can also be used to address the challenge of imbalanced AU data by providing a few examples of positive/negative class from the target tasks, which is typically not costly to obtain. The main contributions of this paper can be summarized as follows:
\begin{itemize}
   \item We propose a new approach for the AU detection where the model adaptation is performed using the notion of model-agnostic meta-learning. Since this approach uses only a few (balanced) shots of the target tasks during the model adaptation, it naturally handles the challenge of model training when faced with highly imbalanced data (as in the case of the AU activations). 
   \item The proposed deep model for the AU detection can easily be adapted to a new subject using only a few shots of target AUs of the target subject. The resulting personalized model for the AU detection largely outperforms the non-personalized model. 
   \item The proposed approach can easily be adapted for detection of AUs that were not used to train the model. This is achieved using only a few shots (the positive/negative examples) of the target AU in order to adapt the model parameters. 
   \item The adaptation to the new task in the proposed approach is very fast, as it does not require a heavy re-training of the deep network since only a few shots are used for adaptation to (subject) or learning of (AUs) the new task.
\end{itemize}

\begin{figure}[t]
\begin{subfigure}{0.5\textwidth}
\centering\includegraphics[width=1\linewidth]{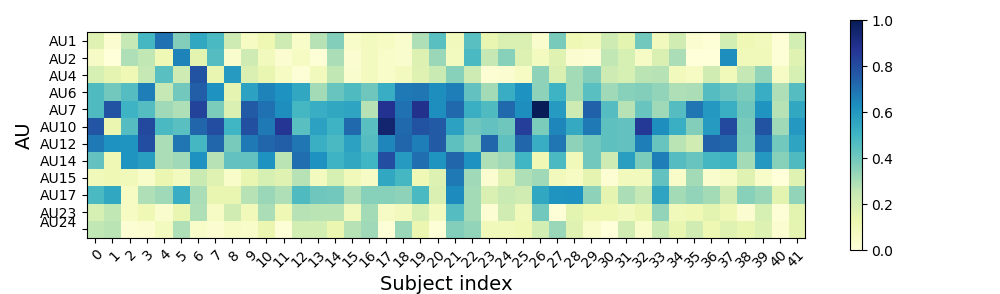}
\caption{BP4D Database}
\label{fig:subim1}
\end{subfigure}
\begin{subfigure}{0.5\textwidth}
\centering\includegraphics[width=1\linewidth]{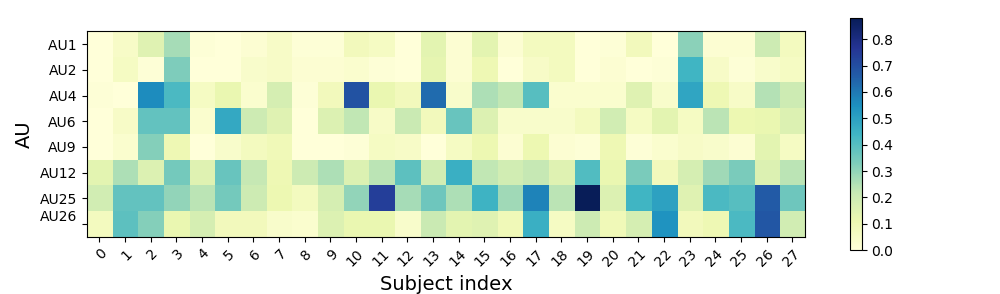}
\caption{DISFA Database}
\label{fig:subim2}
\end{subfigure}
\caption {The portion of the positive examples of each AU and for each subject in the two datasets used in this paper. The evident and highly pronounced imbalance of these data is one of the factors that make the task of learning models for facial AU detection challenging.}
\label{fig:ratio}
\end{figure}

\section{Related Work}
This paper combines a deep facial AU detection model with the few-shot meta-learning for fast model adaptation~\cite{Finn2017ICML}. In this section, we briefly review several relevant works addressing these modeling tasks.
\subsection{Facial Action Unit Detection Model}
Most of the recent works on facial AU detection are based on deep convolutional networks. For instance, Wei et al. \cite{Li2017ActionUD} leveraged an attention mechanism and combined with individual regions of interest (ROIs) cropping as well as the temporal information of expressions with several stacks of the LSTM models. Also, the multi-label learning enabled them to learn relationships between various AUs.~\cite{Zhiwen2018eccv} proposed JAA-Net, which jointly learned both the AU detection and the corresponding best face alignment for the AU, rather than fixing the alignment as a pre-processing step. Also, they defined ROIs of AUs by having AU-specific local and global feature extractors based on hierarchical and  multi-scale region layers with Deep Region and Multi-label Learning (DRML) network proposed by \cite{Zhao2016Deep} . All of these are jointly optimized to facilitate the discovery of the salient local features for the AU detection. Similarly, Wei et al. \cite{Li2017EACNetAR} introduced Enhancing and Cropping (EAC) Net where E-net enhances ROIs of the facial landmark features with attention map and c-net. These are designed for the extraction of "deeper" features using individual CNNs for each facial region through facial regions cropping around the detected landmarks. However, this approach has a limited generalization ability when applied to new subjects or a new AUs, which usually have data distributions different from the training set.

To tackle the challenges of the data distribution shift, several approaches us the models for domain adaptation and transfer learning. Can et al. \cite{Can2018ACM} proposed a generative adversarial recognition network (GARN) which adapts the model from the source subjects to a target subject for personalized AU recognition in unsupervised way. Likewise, Wen et al. \cite{Wen2017tpami} attempted to personalize the facial AU analysis based on SVMs. To match distribution between training and test subjects, \cite{Wen2017tpami} used the re-sampled training data that are most similar in distribution with that of a new test subject, and those are used to re-train the target model in an unsupervised fashion. They alternated between decision function and the selection coefficient by translating the problem into a biconvex problem \cite{Jochen2007biconvex}.
Though both methods address the AU detection problem without any test label, this method requires a lot of time to re-learn the model for a specific test subject. \cite{Can2018ACM} confines their adaptation to the feature extraction level without using any labeled data and training again the feature extractor with a new subject. In~\cite{Zhang2018cvpr}, the authors measure the discrepancy between a new subject and training subjects to find the most similar distribution, and then re-sample the most similar training subject in order to adapt their model to a new subject. 

There have also been a few studies on learning the AU detection model without the AU annotations. For instance, in a recent work, \cite{Zhang2018cvpr} proposed a method that uses prior probabilities on AUs, obtained by leveraging the information about the facial anatomy and emotion expressions. Though they achieved competitive results compared to the supervised methods, their method needs auxiliary information to compensate for the lack of the annotated data. However, this model has a limitation in cases when we have no additional domain knowledge or we need to perform the model adaptation using only a few data examples. 

Different from the approaches mentioned above, this paper investigates the fast model adaptation where the main assumption is that only a few labeled data is available for the target task. Here, a new AU as well as a new subject are regarded as a new task.

\subsection{A Few-Shot Learning}
Few-shot learning is a general modeling framework for fast learning from rarely labeled data. Considering the model is expected to learn and re as fast as possible, designing a model that can be rapidly updated with only few labeled data is essential. However, it is not easy to update the deep neural network with just few labeled samples for a new task since the number of labeled data is insufficient to fine-tune it, which makes the model still over-fitted to training tasks. To address this problem, many few-shot learning and even zero-shot learning studies such as \cite{Lee2017MultilabelZL,Rios2018FewShotAZ,Arora:cvpr} have been introduced. Zero-shot learning does not use any labeled data for a new task but they requires side information such as relationship between seen and unseen task or attribute of each task. Thus, it is also challenging to make such a semantic knowledge as much as to label few samples. 

\begin{figure}[t]
\begin{subfigure}{0.5\textwidth}
\centering\includegraphics[width=0.95\linewidth]{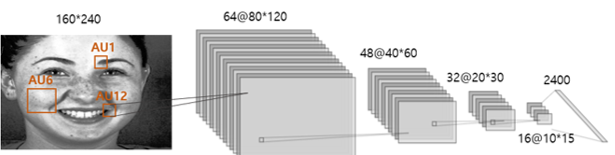}
\caption{Network architecture}
\label{fig:subim1}
\end{subfigure}
\begin{subfigure}{0.5\textwidth}
\centering\includegraphics[width=0.75\linewidth]{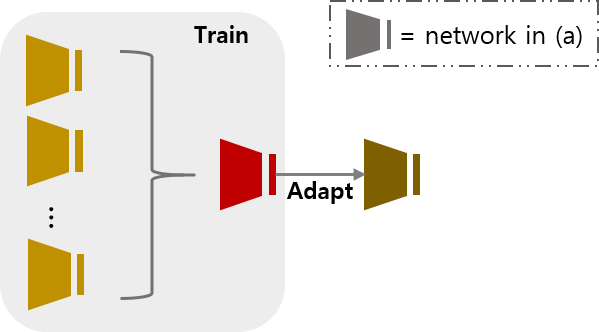}
\caption{Sequence of learning network}
\label{fig:subim2}
\end{subfigure}
\caption {(a) The backbone network used for the AU detection. (b) The sequence of learning the backbone network in (a). Each task-specific model (light yellow network) is trained for each of training tasks according to a subject and AU. Based on per-task trained models, the meta-learned model (red network) is obtained using the discrepancy across tasks. By adapting the meta-learned model to a new task, the new task-specific model (dark yellow network) is achieved. All of these networks have the same architecture as the the backbone network in (a).}
\label{fig:seq_network}
\end{figure}

Model-Agnostic Meta-Learning for Fast Adaptation of Deep Networks (MAML) \cite{Finn2017ICML} introduced a model which is ready to go for fast adaptation with just few labeled data based on gradient descent optimizer. MAML has the meta-learner who learns the differences across training tasks as well as the learner who learns and optimize each of tasks itself. Comparing other few-shot learning methods such as \cite{Ravi2017OptimizationAA} which needs additional network as LSTM to train a classifier or \cite{Koch2015Siamese} which obtains feature embedding with Siamese network and add non-parametric method for a new task, MAML only need only one target network to train meta-learner and the learner as described in Figure \ref{fig:seq_network}. Any model network can be used the one and only backbone network, which makes it called \textit{model-agnostic}.

Recently, \cite{won2019personalized} suggested a personalisation method to estimate the engagement level from videos with only few labeled data. They adopted Active Learning policy from \cite{Woodward2017AL} on their Reinforcement Learning model in order to decide whether keeping estimation or asking new personalized labels from human annotator given a new subject. This paper improves the efficiency for the use of labeled data considering labeling cost.

\section{Methodology}
As a meta-learning approach for the AU detection model, we adopt MAML \cite{Finn2017ICML}. The main idea of MAML is to initialize the weights of deep networks so that it can be updated rapidly for a new task with just a few labeled samples. By adopting MAML, the goal is to find a model that can be adapted quickly to a new task which can be a new subject, a new AU, or both. The Sec. 3.1 describes the network architecture, and the Sec. 3.2 details how we apply the MAML to the AU detection model. Lastly, the Sec. 3.3 describes the baseline approach, i.e., when no benefit of meta-learning is made for adaptation.

\algdef{SE}[SUBALG]{Indent}{EndIndent}{}{\algorithmicend\ }%
\algtext*{Indent}
\algtext*{EndIndent}

\begin{algorithm}[t]
\small
\caption{MAML: Training phase}
\label{alg}
\begin{algorithmic}[1]
\State \textbf{Input}: $p(T)$, distribution over tasks
\State \textbf{Require}: $\alpha, \beta$ step size hyper parameters
\State \textbf{Init}: $\theta$ randomly
\Repeat
\State \text Sample tasks ${T_i}$ ~ $p(T)$ for i= 1, \dots, $|T|$
\State \text Create few-shot sample set $D_{T_i}= D_{T_i}[tr] \cup D_{T_i}[trv]$ for i= 1, \dots, $|T|$
\State \textbf{for i=1,\dots,$|T|$}
\Indent
\State \text Evaluate $L_{T_i}$ using $\theta$ and $D_{T_i}[tr]$
\State \text Update $\theta _i = \theta - {\alpha}{\nabla _{\theta} L_{T_i}}$

\State \text  Evaluate $L_{T_i}$ using updated $\theta _i$ and  $D_{T_i}[trv]$
\EndIndent
\State \textbf{end for}
\State ${\theta \leftarrow \theta - \beta{\nabla _{\theta}}\sum\limits_{i = 1}^{|T|} L_{T_i}}$

\Until {convergence}
\State \textbf{Output}: $\theta$
\end{algorithmic}
\label{alg:my_algo}
\end{algorithm}

\subsection{Model Network}
Before passing through our own network, input images are pre-processed with Deepfakes~\cite{deepfake} to extract and crop the face region. The proposed deep network applied to the pre-processed face images is illustrated in Figure \ref{fig:seq_network} (a). The encoder consists of 4 modules with 64, 48, 32, and 16 filters of 5 convolutions, respectively, followed by the batch normalization, ReLU, and max-pooling layers. This is further followed by a fully-connected (FC) layer and a sigmoid function, used to train the detectors for the target AU. Note that the proposed detector is designed for detecting only one AU at the time, which is consequently adapted for the detection of a new target AU using the meta-learning as explained below.

\begin{figure*}[t]
\begin{center}
\begin{subfigure}[t]{0.52\linewidth}
\centering\includegraphics[width=\linewidth]{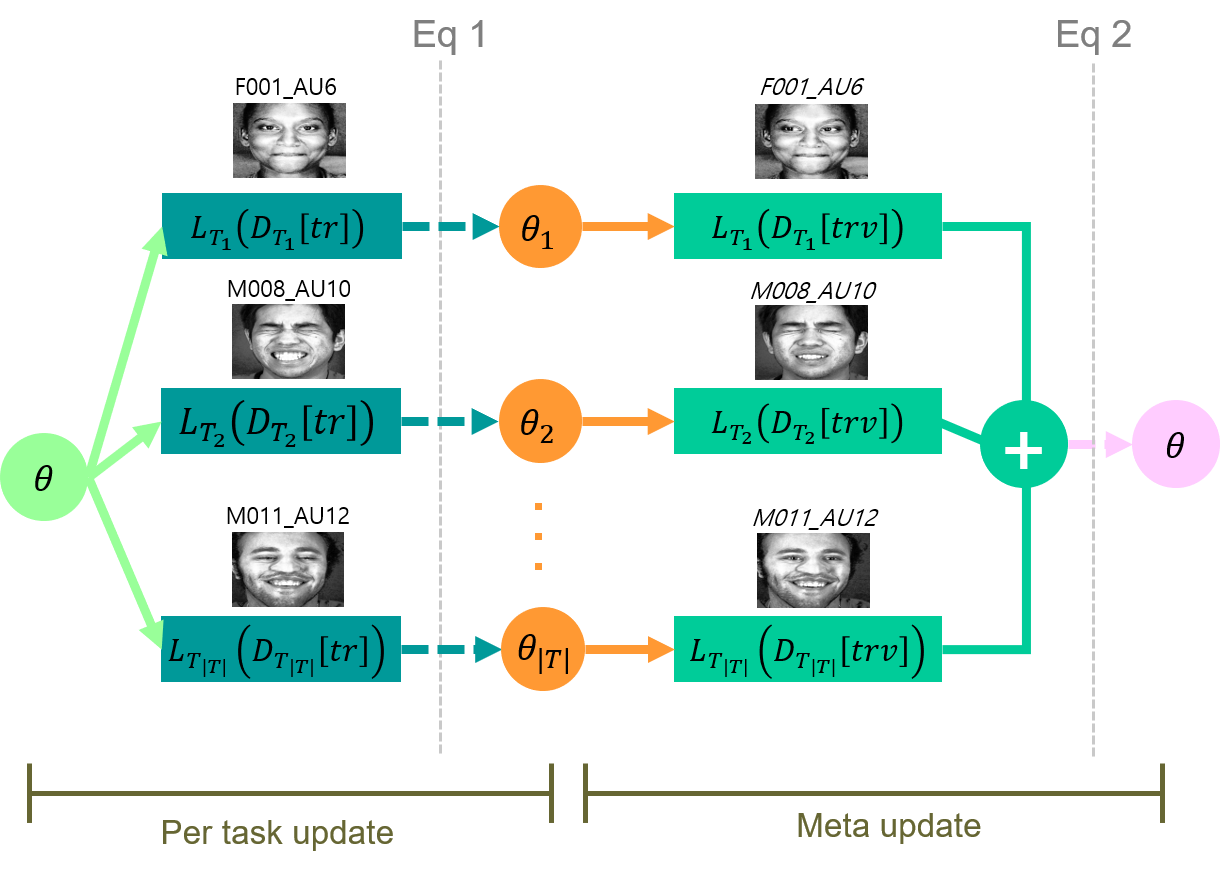}
\caption{Meta-learning model in training phase.}
\end{subfigure}\hfill
\begin{subfigure}[t]{0.44\linewidth}
\centering\includegraphics[width=\linewidth]{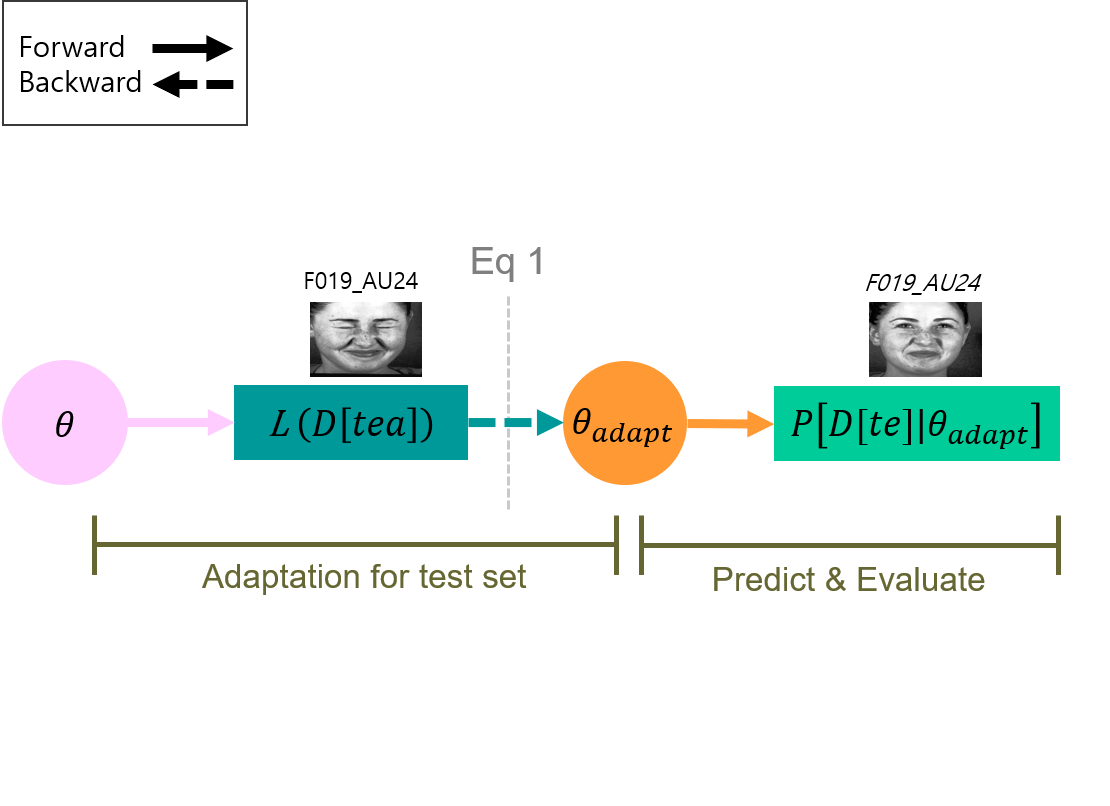}
\caption{Adaptation in test phase. }
\end{subfigure}

\end{center}
  \caption{(a) The per-task learning and meta-learning during the model training phase. Any subject and AU can be consisdered a task except for one subject in our leave-one-subject-out setting (for example, in this figure, the F019 from the BP4D dataset). The same tasks but different instances are used in each of \textit{per-task update} and \textit{meta-update}. After the meta-parameter $\theta$ is updated as the per-task $\theta_i$ for each task, each of these are used to compute the second loss from $D_{T_i}[trv]$. The losses are summed up across all tasks to obtain the new meta-updated parameter $\theta$. (b) The adaptation phase. In our leave-one-subject-out setting, we leave all AUs of the one left subject as test tasks. (b) An example of AU24 of F019 -- after the few-shot learning of the new given task, our model adapt its meta-parameter best fitted for that task so that it can make better predictions of the target AU.}
\label{fig:arch}
\end{figure*}

\subsection{Meta-Learned Facial AU Detection Model}

Model-Agnostic Meta-Learning(MAML) \cite{Finn2017ICML} is gradient-based meta-learning for multiple different tasks with few-shot learning. Figure \ref{fig:seq_network} (b) illustrates the overall sequence of training our model based on MAML. Each task-specific model (light yellow network) is trained for each of training tasks, which corresponds to Alg.~\ref{alg:my_algo} (lines 7--11). Those per-task trained parameters are aggregated to train the meta-learned model which is the red network. This whole meta-learning procedure is from line 4--12 in Alg.~\ref{alg:my_algo}. After having the optimal meta-learned model (red network), by adapting it to a new task, the new task-specific model (dark yellow network) is obtained. More detailed training and adaptation architecture can be found in Figure \ref{fig:arch}.
Though we used a simple network depicted in Figure \ref{fig:seq_network} (a), any model can be assembled in Figure \ref{fig:seq_network} (b) sequence which is model-agnostic. In the following, we explain in more detail how the per-task learning and meta-learning are implemented.

\subsubsection{Per-task Learning}
Per-task learning can be seen as the traditional learning approach: choosing a backbone model for the target dataset, and training the model to be best fitted for each training task of the dataset. The difference with other methods is that we divide our training set into tasks according to a subject and AU, and train each of them separately as shown in Figure \ref{fig:arch} (a) whereas other methods use the whole training set to train their model. Alg. \ref{alg:my_algo} describes per-task learning in line 7 to 11. Given the objective function ${L}$, the per-task loss is computed using $D_{T_i}[tr]$, which denotes a batch of training data for task ${T_i}$. Based on the obtained loss, the initial weight $\theta$ is updated into $\theta_i$, which is the optimal learned weight for the target task ${T_i}$. 
\begin{equation}
    \theta _i = \theta - {\alpha}{\nabla _{\theta} L_{T_i}}
\label{eq:per_task}
\end{equation}

\subsubsection{Meta-learning}
Our goal is to learn the optimal meta-weight that can quickly be adapted to a new task in test time. To do so, in meta-learning phase, the model learns the discrepancy across all learned per-task models.
Based on every optimal $\theta_i$ from per-task learning, the loss is re-computed. This is shown on line 10 of Alg. \ref{alg:my_algo} and illustrated in \textit{Meta-update} part of Figure \ref{fig:arch} (a). To achieve this, we need another set of training data of each task, which is a disjoint set $D_{T_i}[tr]$ so that this another set can play a role of validation set for the new loss. It should be noted that this validation set is still from the same training task. We denote $D_{T_i}[trv]$ as training-validation set in each training task ${T_i}$. The re-computed loss across all tasks is summed up and based on this total loss, we obtain the meta-updated $\theta$.  
\begin{equation}
    {\theta \leftarrow \theta - \beta{\nabla _{\theta}}\sum\limits_{i = 1}^{|T|} L_{T_i}}
\label{eq:meta}
\end{equation}

Note that the goal of the training phase is to output $\theta$ that can be fast adapted to a new task, not to output $\theta_{i}$ for each of training tasks ${T_i}$. In adaptation phase, initialized by the trained $\theta$, the model can reach to the optimal $\theta_{j}$ quickly for a new task ${T_j}$.

In summary, the model minimizes the discrepancy across different tasks with meta-learning as well as the individual loss for each task with per-task learning. Our model defines each AU and subject as a task. Thus, in training phase, it learns how to adapt fast across the different subjects and AUs. After then, given a new task with few labeled data, it can be rapidly customized to the new task. One notable part is that we have only one classifier for any AU as well as any subject. By having this architecture, 1) we can increase the number of tasks to learn the discrepancy, 2) the task distribution can have higher variance which makes the model learn meaningful diversity in terms of meta-learning, and 3) our model can be adapted fast both to new AU and new subject. The objective function $L_{T_i}$ in Eq. \ref{eq:per_task} and \ref{eq:meta} is cross-entropy loss for binary classification in our model.
\begin{equation}
    L_{T_i} = y_{T_i} \log\sigma(x_{T_i};\theta_{T_i}) + (1-y_{T_i}) (1-\log\sigma(x_{T_i};\theta_{T_i}))
\label{eq:loss}
\end{equation}

\subsection{Baseline}
To verify that the rapid and effective adaptation to a new task comes from the meta-learned knowledge, we need a baseline which has no ability of meta-perception. To make such a model, \textit{multi}-tasks concept should be excluded in meta-update phase. In other words, the same network as ours but standard deep learning process with only one task. More importantly, since our model has only one classifier which can be adapted to any AU as well as any subject, the baseline should be also a model with just one classifier for all available tasks. However, existing approaches use a customized classifier for each AU, which makes them improper as the baseline for our adaptation setting. Thus, we build our own baseline model with one classifier for the whole training set without the task separation. We set a label as \textit{negative} if none of training AUs is detected in the given sample, and \textit{positive} otherwise. In other words, a sample is labeled as positive as long as at least one of AU is detected. After training, the same adaptation methodology as in our model is applied to the baseline for comparison.

\section{Experiments}

\subsection{Datasets and Tasks}
We have evaluated our model on two benchmark facial AU datasets - DISFA \cite{Mavadati2013IEEE} and BP4D \cite{Mavadati2013IEEE}.
\subsubsection{BP4D} This dataset  consists of data of 41 subjects (23 females and 18 males). For each subject, there are 2-D and 3-D videos (we use only the former) of 8 sessions, where some of frames are positive or negative annotation per AU. The total number of valid frames with label is 140K. As in \cite{Zhiwen2018eccv}, \cite{Li2017ActionUD}, \cite{Li2017EACNetAR}, we also limited the number of AUs to 12 AUs consisting of AU1, AU2, AU4, AU6, AU7, AU10, AU12, AU14, AU15, AU17, AU23 and AU24.
\subsubsection{DISFA} This dataset consists of data of 27 subjects (12 females and 15 males). Each subject has a video of 4,845 frames where each of the frames has 0--5 intensity per AU. As in \cite{Zhiwen2018eccv}, \cite{Li2017ActionUD}, \cite{Li2017EACNetAR}, we also limited the number of AUs as 8 AUs consisting of AU1, AU2, AU4, AU6, AU9, AU12, AU25 and AU26. Since our model needs the balanced positive and negative sample ratio, in order to maximize positive labeled samples, any intensity larger than 0 is regarded as positive. We have conducted two experiments on DISFA. One is the standard setting as training with training set of DISFA followed by adapting and evaluating with test set of DISFA. Another is training with whole BP4D and then adapt and evaluate with DISFA in order to see how a new AU is rapidly learned in our model as well as to see more distinct subject-adaptation result. 
\subsubsection{Task}
In each dataset, combinations of the above elaborated subjects and AUs are regarded as a task in our model. For learning, it is crucial to have as many tasks as possible considering the meta-knowledge comes from learning discrepancy across different tasks. Thus, as in \cite{Can2018ACM}, by adopting leave-one-subject-out cross validation, we increased the number of tasks. Thus, in BP4D, we have 12 AUs $\times$ 39 subjects = 468 different tasks in training. In DISFA, 8 AUs $\times$ 26 = 208 tasks. And the last subject's all AUs are adapted and evaluated per-task.

As described in Sec. 3.2, there are two learning stages: 1) training with sets $D_T[tr]$ and $D_T[trv]$, and 2)adaptation with sets $D_T[te]$ and $D_T[tea]$, where $D_T[tea]$ denotes the labeled test data used to adapt to the test task, and $D_T[te]$ denotes unlabeled data used for the model evaluation. 

\subsubsection{Per-task sampling}
\noindent When sampling a set of batch data, it is crucial to keep balanced positive and negative pairs of samples in order to prevent biased learning, adaptation, and evaluation. However, as seen in Figure \ref{fig:ratio} (a), negative samples are dominant. To address this issue, we repeatedly used the positive samples with small batch size that comes from few-shot learning. During training, we need to select balanced N-positive and N-negative samples, denoted by (N+, N-). It is required to have one set of (N+, N-) for $D[tr]$ and another exclusive set of (N+, N-) for $D[trv]$ per one task, total 4N labeled samples. We set $N = 5$ for training no matter how many few-shot learning we do during test phase since we can assume and indeed have enough labeled training data.

On the other hand, in adaptation phase, we only need one set of labeled samples (K+, K-) for $D[tea]$ since we do not have to do meta-update but only per-task update is required. Here, K varies depending on how many labeled data we can have for the given new task. Unlabeled test data $D[te]$ is used to evaluate the performance after adaptation.

\subsubsection{Balancing samples per-task}
\noindent For some AUs, there are very few positive labels, which makes it difficult to create balanced pair when it is less than N or K. In such a case, we can think of two strategies - skip that task or replace the insufficient portion of positive samples with negative samples. For training phase, we adopt skipping the task. This is because 1) under the $K = 5$ shot training, we miss only two tasks in BP4D and 13 tasks in DISFA from the total of 468 and 208 total tasks respectively, and 2) those skipped tasks can be learned and covered in some degree from another AU or subject. For instance, assume a task consisting of AU1 and subject 1 is omitted. Though AU1 is missed out of subject 1, our model can still exploit the data of other subjects' AU1 properties. Also, the subject 1's personalized features could be achieved from other tasks (subject 1 but other AUs combination). In test time, however, the lack of positive samples are filled with negative samples since there is no other task but the target task. For example, each of $D[tea]$ and $D[te]$ have ((K-m)+, (K+m)-) pair when only K-m, which is less than K, positive samples are available.

\begin{table}[t]
\centering
\begin{tabular}{lrrrr}  
\toprule

\multirow{2}{*} {AU} &
  \multicolumn{2}{c}{1shot} &
  \multicolumn{2}{c}{5shot} \\ 
 & base & ours  & base & ours \\
\midrule
AU1&0.57&0.6&0.59&0.71\\
AU2&0.59&0.61&0.62&0.72\\
AU4&0.58&0.61&0.63&0.74\\
AU6&0.71&0.71&0.74&0.81\\
AU7&0.67&0.67&0.69&0.75\\
AU10&0.72&0.73&0.75&0.83\\
AU12&0.75&0.77&0.78&0.86\\
AU14&0.61&0.61&0.61&0.7\\
AU15&0.59&0.64&0.6&0.73\\
AU17&0.56&0.62&0.56&0.71\\
AU23&0.56&0.62&0.57&0.72\\
AU24&0.56&0.67&0.58&0.8\\
AVG&0.62&{\bf0.66}&0.64&{\bf0.76}\\
\bottomrule
\end{tabular}
\caption{Averaged accuracy across subjects reported per AU on {\bf BP4D}. The results are after the adaptation with 5 gradient steps.}
\label{tab:BP4D}
\end{table}

\begin{table}[t]
\centering
\begin{tabular}{lrrrr}  
\toprule
\multirow{2}{*} {AU} &
  \multicolumn{2}{c}{1shot} &
  \multicolumn{2}{c}{5shot} \\ 
 & base & ours  & base & ours \\
\midrule
AU1&0.64&0.68&0.66&0.74\\
AU2&0.69&0.74&0.7&0.8\\
AU4&0.61&0.65&0.62&0.72\\
AU6&0.77&0.8&0.8&0.87\\
AU9&0.77&0.81&0.8&0.88\\
AU12&0.72&0.76&0.75&0.83\\
AU25&0.73&0.73&0.75&0.79\\
AU26&0.68&0.69&0.71&0.75\\
AVG&0.7&{\bf0.73}&0.72&{\bf0.8}\\
\bottomrule
\end{tabular}
\caption{Averaged accuracy across subjects reported per AU on {\bf DISFA}. The results are after the adaptation with 5 gradient steps.}
\label{tab:DISFA}
\end{table}

\subsection{Experimental Setting}
\subsubsection{Implementation Details}
When we apply meta-update to compute $\theta$, we need to aggregate all different tasks, where the number of tasks should be concerned. We chose 12 tasks for BP4D and 8 tasks for DISFA. This number is set according to the total number of AUs in each dataset, as described in Sec. 4.1. We implemented our experiments with $K = 1$, or $5$ shot learning. In other words, we adapt our trained model with 1-shot (1+, 1-) or 5-shot (5+,5-) samples of a new task in test time, as these are easy to obtain in the real-world settings. Adam optimizer was used with learning rate 0.03 for both $\alpha$ and $\beta$ in Alg. \ref{alg:my_algo}
\subsubsection{Evaluation Method}
As described in Sec. 4.2, a set of few-shot samples $D[tea]$ is used for adaptation. After that, we evaluate the adapted model with another balanced set $D[te]$, which consists of 10 positive and 10 negative samples (used only for the model evaluation). We repeat this K-shot adaptation and 10-shot evaluation for 500 times and average all results. If $D[te]$ cannot be formulated in completely balanced way because of lack of positive samples, we replace the deficient positive samples with negative samples as described in Sec. 4.2. Since we use a balanced set for evaluation, we report accuracy as our evaluation metric.

\begin{figure}[t]
\begin{subfigure}{0.5\textwidth}
\includegraphics[width=0.95\linewidth]{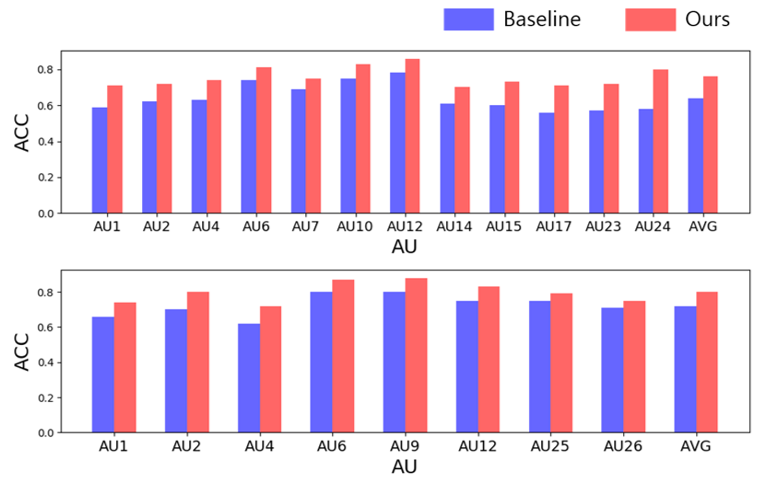}
\caption{Averaged ACC across subjects per AU}
\label{fig:subim1}
\end{subfigure}
\begin{subfigure}{0.5\textwidth}
\includegraphics[width=0.95\linewidth]{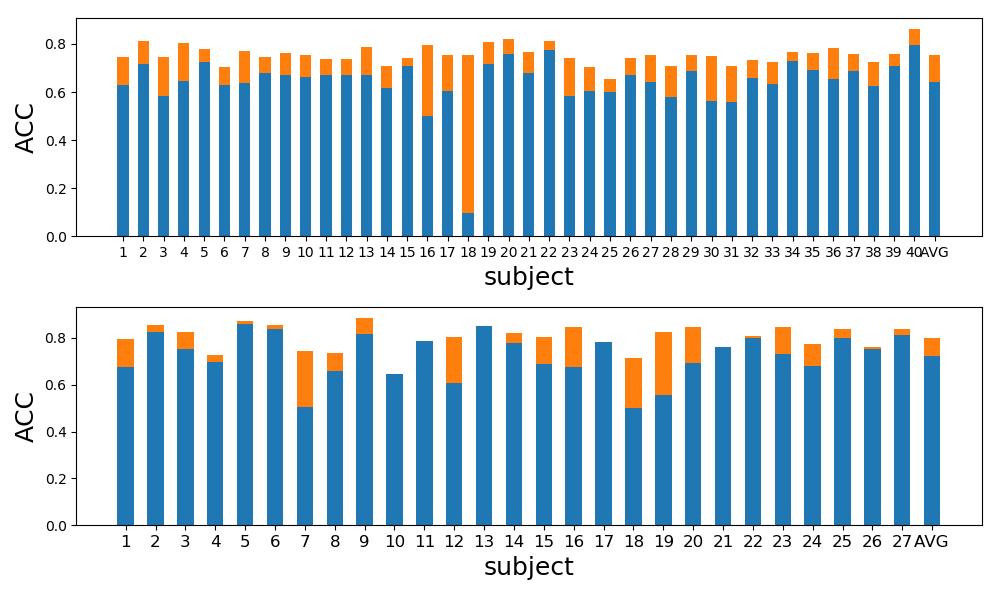}
\caption{Averaged ACC across AUs per subject}
\label{fig:subim2}
\end{subfigure}
\caption {Analysis of the adaptation for the BP4D and DISFA datasets. For both (a) and (b), results are after adaptation with 5-shot and 5 gradient steps. (a) shows the averaged accuracy across all subjects per AU (BP4D at the top, and DISFA at the bottom). The proposed approach (ours) achieves better accuracy than the baseline on all AUs, as expected. (b) The averaged accuracy across all AUs per subject in each dataset (BP4D at the top, and DISFA at the bottom). The averaged accuracy of the baseline is illustrated by the blue bar. The yellow bar above the blue bar indicates the increase of the averaged accuracy by using our model.}
\label{fig:fera_disfa}
\end{figure}

\subsection{Adaptation to New Subjects - Personalisation}
For a new subject as a new task, while all AUs are used during training, we can regard our model as a personalized model since it is adapted and optimized for the newly given subject. The performance on BP4D dataset is shown in Table \ref{tab:BP4D}. During training, all BP4D subjects except for one subject is trained. And then in adaptation, the one left BP4D subject is tested. After this leave-one-subject-out experiment is done for every subject, results are averaged across all the subjects per each AU. Table \ref{tab:DISFA} shows the result of DISFA dataset with the same protocol as in BP4D. The corresponding graph of each dataset can be checked in Figure \ref{fig:fera_disfa} (a). On average, our model outperforms the baseline in both 1-shot and 5-shot learning for each of two datasets, as expected. One notable part is that by increasing the number of test samples from 1-shot to 5-shot, the gap between ours and baseline becomes more significant. In other words, as more samples are given, the more boosted meta-learned effects we can have. It can be interpreted as our model does know better than baseline on how well utilizing given labeled samples of a new task through the meta-learning knowledge. The same experiments but the result averaged across all AUs are shown in Figure \ref{fig:fera_disfa} (b). For majority of subjects, our model achieves higher accuracy on averaged AU detection performance.

\begin{table}[t]
\centering
\begin{tabular}{lrrrr}  
\toprule
\multirow{2}{*} {AU} &
  \multicolumn{2}{c}{1shot} &
  \multicolumn{2}{c}{5shot} \\ 
 & base & ours  & base & ours \\
\midrule
AU1&0.64&0.67&0.64&0.77\\
AU2&0.69&0.74&0.71&0.82\\
AU4&0.59&0.65&0.58&0.75\\
AU6&0.78&0.79&0.82&0.87\\
AU9&0.75&0.79&0.78&0.87\\
AU12&0.74&0.75&0.77&0.82\\
AU25&0.67&0.67&0.68&0.75\\
AU26&0.64&0.65&0.65&0.73\\
AVG&0.69&{\bf0.71}&0.7&{\bf0.8}\\
\bottomrule
\end{tabular}
\caption{Averaged accuracy per AU across all leave-out-out test subjects on {\bf DISFA}. Results are after adaptation with 5 gradient steps from the {\bf BP4D-trained} model. }
\label{tab:BP4D_DISFA}
\end{table}

\begin{figure}[t]
\begin{subfigure}{0.5\textwidth}
\begin{center}
\includegraphics[width=0.95\linewidth]{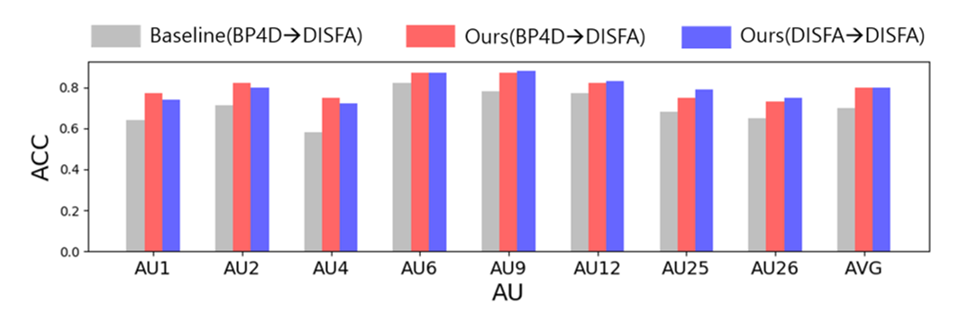}
\end{center}
\caption{Averaged ACC across subjects per AU}
\label{fig:subim1}
\end{subfigure}

\begin{subfigure}{0.5\textwidth}
\begin{center}
\includegraphics[width=0.65\linewidth]{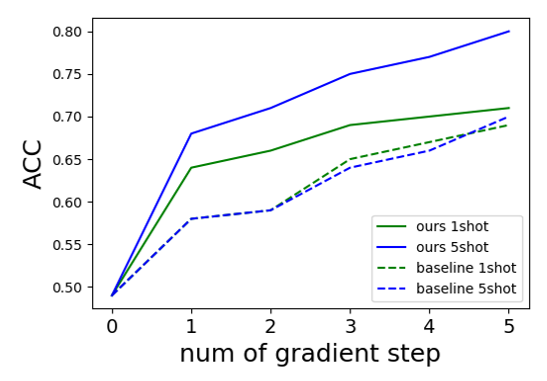}
\end{center}
\caption{ACC change according to the number of gradient step}
\label{fig:subim2}
\end{subfigure}
\caption {Analysis on domain transfer from BP4D to DISFA. In (a), the accuracy [in \%] is averaged across subjects per AU, adapted and evaluated on DISFA from three different trained models; BP4D baseline (grey), our BP4D (red), and our DISFA (blue). (b) illustrates how accuracy changes according to the number of the gradient steps. The accuracy is averaged across AUs/subjects. }
\label{fig:fera_to_disfa}

\end{figure}

\subsection{Adaptation to New AUs - Domain Adaptation}
Even a completely new AU can quickly be learned with our meta-trained classifier since we treat an AU as a task in the training. We verified this by training the model with BP4D and adapting it with DISFA. AU9, 25, and 26, which exist only in DISFA not in BP4D, are used as a new AU in test time. As well as the performance of adaptation on new AU, this experiment can show how well our model can work for domain adaptation since we transfer our dataset from BP4D in training time into DISFA during test time. Both baseline and ours are trained with BP4D and then, 1 or 5-shot DISFA data is injected for adaptation. Since the whole DISFA dataset can be seen as a new task, all tasks from BP4D can be used during training. In other words, no leave-one-subject-out cross validation is needed as in \textit{Adaptation to New Subjects}. Table \ref{tab:BP4D_DISFA} reports that ours outperforms baseline with the same pattern as in \textit{Adaptation to New Subjects} where 5-shot shows more substantial difference with baseline than 1-shot. 

To check the effects of domain adaptation (in our case, the cross-dataset adaptation), Figure \ref{fig:fera_to_disfa} (a) illustrates comparison with DISFA-trained model of which training and test subject is from DISFA under leave-one-subject-out cross validation. For AU1, 2, and 4 which are co-existing AUs in both of datasets, our BP4D-trained model shows better accuracy than our DISFA-trained model. This is because BP4D has more training tasks than DISFA, from which our model can learn more meta-information in order to adapt well. For AU9, 25, and 26, which are new AUs to BP4D, our BP4D-trained model can achieve almost the same performance as our DISFA-trained model. In conclusion, considering average accuracy of our BP4D is same as that of our DISFA, our model outperforms in few-shot domain transfer learning. For some common AUs (AU1, 2, and 4), our BP4D can almost beat our DISFA for completely previously unseen labels of AUs (AU9, AU25, and AU26), which is an evidence of creating one powerful base classifier to make a baseline prediction for any AU.

Figure \ref{fig:fera_to_disfa} (b) shows the BP4D-trained model and BP4D-trained baseline's improvements in relation to the number of gradient step for each of 1 and 5 shot adaptation. Our approach accomplishes larger improvement than the baseline with both few-shot adaptations. Also, our model shows better accuracy overall in 5 shot rather than in 1 shot whereas baseline has almost no difference between 1 shot and 5-shot. This implies that our meta-learning knowledge makes it possible to leverage few-shot samples more efficiently for adaptation than the baseline. In addition, with just one gradient step, our model shows steep increase in accuracy though it can keep improving with more steps. This evidences that our model is trained in a way that enables to adapt very fast to a new task.

\section{Conclusion}
In this paper, we introduced a meta-learning approach for facial AU detection that can quickly adapt to new subjects or AUs. The main idea of this approach is that by optimizing the model prepared for adaptation rather than optimizing it fitted on specific task. We showed that this approach can fully utilize a few labeled data of a new task to learn the new task as well as largely improve the performance by just one gradient update. In the future work, we will devise appropriate evaluation way to make comparison with other related works on domain adaptation and transfer learning, applicable to the task of AU detection. 
\section*{Acknowledgements}
The work of O. Rudovic was funded by EC H2020 no. 701236 Marie Curie Action -- Individual Fellowship (EngageMe).

\bibliographystyle{named}
\bibliography{ijcai19}

\begin{thebibliography}{}

\bibitem[\protect\citeauthoryear{Arora \bgroup \em et al.\egroup
  }{2017}]{Arora:cvpr}
Gundeep Arora, Vinay Verma, Ashish Mishra, and Piyush Rai.
\newblock Generalized zero-shot learning via synthesized examples.
\newblock In {\em Proceedings of IEEE Conference on CVPR}, pages 4281--4289, 12
  2017.

\bibitem[\protect\citeauthoryear{C.~Finn and Levine}{2017}]{Finn2017ICML}
P.~Abbeel C.~Finn and S.~Levine.
\newblock Model-agnostic meta learning for fast adaptation of deep networks.
\newblock In {\em Proceedings of ICML}, pages 1126--1135, 2017.

\bibitem[\protect\citeauthoryear{Chu \bgroup \em et al.\egroup
  }{2016}]{Wen2017tpami}
Wen-Sheng Chu, Fernando De~la Torre, and Jeffrey Cohn.
\newblock Selective transfer machine for personalized facial expression
  analysis.
\newblock {\em IEEE Transactions on Pattern Analysis and Machine Intelligence},
  39(3):529--545, 03 2016.

\bibitem[\protect\citeauthoryear{Deepfakes}{}]{deepfake}
Deepfakes.
\newblock https://github.com/deepfakes/faceswap.

\bibitem[\protect\citeauthoryear{Gorski \bgroup \em et al.\egroup
  }{2007}]{Jochen2007biconvex}
Jochen Gorski, Frank Pfeuffer, and Kathrin Klamroth.
\newblock Biconvex sets and optimization with biconvex functions: A survey and
  extensions.
\newblock {\em Mathematical Methods of Operations Research}, 66:373--407, 11
  2007.

\bibitem[\protect\citeauthoryear{Koch \bgroup \em et al.\egroup
  }{2015}]{Koch2015Siamese}
Gregory Koch, Richard Zemel, and Ruslan Salakhutdinov.
\newblock Siamese neural networks for one-shot image recognition.
\newblock In {\em ICML deep learning workshop}, 2015.

\bibitem[\protect\citeauthoryear{Lee \bgroup \em et al.\egroup
  }{2017}]{Lee2017MultilabelZL}
Chung-Wei Lee, Wei Fang, Jason Yeh, and Yu-Chiang~Frank Wang.
\newblock Multi-label zero-shot learning with structured knowledge graphs.
\newblock In {\em Proceedings of IEEE Conference on CVPR}, pages 1576--1585,
  2017.

\bibitem[\protect\citeauthoryear{Li \bgroup \em et al.\egroup
  }{2017a}]{Li2017ActionUD}
Wei Li, Farnaz Abtahi, and Zhigang Zhu.
\newblock Action unit detection with region adaptation, multi-labeling learning
  and optimal temporal fusing.
\newblock In {\em Proceedings of IEEE Conference on CVPR}, pages 6766--6775,
  2017.

\bibitem[\protect\citeauthoryear{Li \bgroup \em et al.\egroup
  }{2017b}]{Li2017EACNetAR}
Wei Li, Farnaz Abtahi, Zhigang Zhu, and Lijun Yin.
\newblock Eac-net: A region-based deep enhancing and cropping approach for
  facial action unit detection.
\newblock In {\em Proceedings of IEEE FG}, pages 103--110, 2017.

\bibitem[\protect\citeauthoryear{Lucey \bgroup \em et al.\egroup
  }{2010}]{Lucey2010IEEE}
Patrick Lucey, Jeffrey Cohn, Takeo Kanade, Jason Saragih, Zara Ambadar, and
  Iain Matthews.
\newblock The extended cohn-kanade dataset (ck+): A complete dataset for action
  unit and emotion-specified expression.
\newblock In {\em Proceedings of IEEE Conference on CVPR Workshops}, pages 94
  -- 101, 07 2010.

\bibitem[\protect\citeauthoryear{Mavadati \bgroup \em et al.\egroup
  }{2013}]{Mavadati2013IEEE}
Seyedmohammad Mavadati, Mohammad Mahoor, Kevin Bartlett, Philip Trinh, and
  Jeffrey Cohn.
\newblock Disfa: A spontaneous facial action intensity database.
\newblock {\em IEEE Transactions on Affective Computing}, 4(2):151--160, 2013.

\bibitem[\protect\citeauthoryear{Pantic \bgroup \em et al.\egroup
  }{2005}]{Pantic2005IEEE}
M~Pantic, Michel Valstar, R~Rademaker, and L~Maat.
\newblock Web-based database for facial expression analysis.
\newblock In {\em Proceedings of IEEE ICME}, pages 5 pp.--, 07 2005.

\bibitem[\protect\citeauthoryear{Ravi and
  Larochelle}{2017}]{Ravi2017OptimizationAA}
Sachin Ravi and Hugo Larochelle.
\newblock Optimization as a model for few-shot learning.
\newblock In {\em ICLR}, 2017.

\bibitem[\protect\citeauthoryear{Rios and Kavuluru}{2018}]{Rios2018FewShotAZ}
Anthony Rios and Ramakanth Kavuluru.
\newblock Few-shot and zero-shot multi-label learning for structured label
  spaces.
\newblock In {\em Proceedings of the Conference on Empirical Methods in Natural
  Language Processing}, pages 3132--3142, 2018.

\bibitem[\protect\citeauthoryear{Rudovic \bgroup \em et al.\egroup
  }{2019}]{won2019personalized}
Ognjen Rudovic, Hae Won~Park, John Busche, Bjorn Schuller, Cynthia Breazeal,
  Rosalind~W Picard, et~al.
\newblock Personalized estimation of engagement from videos using active
  learning with deep reinforcement learning.
\newblock In {\em Proceedings of IEEE Conference on CVPR Workshops}, 2019.

\bibitem[\protect\citeauthoryear{Shao \bgroup \em et al.\egroup
  }{2018}]{Zhiwen2018eccv}
Zhiwen Shao, Zhilei Liu, Jianfei Cai, and Lizhuang Ma.
\newblock Deep adaptive attention for joint facial action unit detection and
  face alignment.
\newblock In {\em ECCV}, 2018.

\bibitem[\protect\citeauthoryear{Wang and Wang}{2018}]{Can2018ACM}
Can Wang and Shangfei Wang.
\newblock Personalized multiple facial action unit recognition through
  generative adversarial recognition network.
\newblock In {\em Proceedings of ACM International Conference on Multimedia},
  pages 302--310, 10 2018.

\bibitem[\protect\citeauthoryear{Woodward and Finn}{2016}]{Woodward2017AL}
Mark Woodward and Chelsea Finn.
\newblock Active one-shot learning.
\newblock In {\em NIPS, Deep Reinforcement Learning Workshop}, 2016.

\bibitem[\protect\citeauthoryear{Zhang \bgroup \em et al.\egroup
  }{2018}]{Zhang2018cvpr}
Yong Zhang, Weiming Dong, Bao-Gang Hu, and Qiang Ji.
\newblock Classifier learning with prior probabilities for facial action unit
  recognition.
\newblock In {\em Proceedings of IEEE Conference on CVPR}, pages 5108--5116, 06
  2018.

\bibitem[\protect\citeauthoryear{Zhao \bgroup \em et al.\egroup
  }{2016}]{Zhao2016Deep}
Kaili Zhao, Wen-Sheng Chu, and Honggang Zhang.
\newblock Deep region and multi-label learning for facial action unit
  detection.
\newblock In {\em Proceedings of IEEE Conference on CVPR}, pages 3391--3399, 06
  2016.

\end{thebibliography}
\end{document}